\documentclass[conference]{IEEEtran}
\IEEEoverridecommandlockouts
\usepackage{cite}
\usepackage{amsmath,amssymb,amsfonts}
\usepackage{algorithmic}
\usepackage{graphicx}
\usepackage{textcomp}
\usepackage{xcolor}
\def\BibTeX{{\rm B\kern-.05em{\sc i\kern-.025em b}\kern-.08em
		T\kern-.1667em\lower.7ex\hbox{E}\kern-.125emX}}

\usepackage{graphicx}
\usepackage{amsmath}
\usepackage{algorithm}
\usepackage{paralist}
\usepackage{framed}
\usepackage{amsfonts}
\usepackage{multirow}
\usepackage{courier}
\usepackage{subcaption}
\usepackage{xcolor}
\definecolor{code-blue}{RGB}{8,0,255}
\definecolor{code-red}{RGB}{255,83,112}
\definecolor{code-purple}{RGB}{102,0,153}
\usepackage[american]{babel}
\usepackage{microtype}

\begin{document}
	\title{Improved Dynamic Memory Network for  Dialogue Act Classification with Adversarial Training}
	\author{\IEEEauthorblockN{Yao Wan\textsuperscript{*}$^\text{\dag}$, Wenqiang Yan\textsuperscript{*}$^\dagger$\thanks{\textsuperscript{*}Equal contribution}, Jianwei Gao$^\dagger$, Zhou Zhao$^\dagger$, Jian Wu$^\text{\dag}$, Philip S. Yu$^{\dagger\sharp}$}
		\IEEEauthorblockA{$^\dagger$College of Computer Science and Technology, Zhejiang University, Hangzhou, China\\
			$^\ddagger$Department of Computer Science, University of Illinois at Chicago, Illinois, USA\\
			$^\sharp$Institute for Data Science, Tsinghua University, Beijing, China\\
			\{wanyao,gjwei,zhaozhou,wujian2000\}@zju.edu.cn, yxlink2017@gmail.com, psyu@uic.edu}
	}
	\maketitle
	
	\begin{abstract}
		Dialogue Act (DA) classification is a challenging problem in dialogue interpretation, which aims to attach semantic labels to utterances and characterize the speaker's intention.
		Currently, many existing approaches formulate the DA classification problem ranging from multi-classification to structured prediction, which suffer from two limitations: a) these methods are either handcrafted feature-based or have limited memories. b) adversarial examples can't be correctly classified by traditional training methods. To address these issues, in this paper we first cast the problem into a question and answering problem and proposed an improved dynamic memory networks with hierarchical pyramidal utterance encoder. Moreover, we apply adversarial training to train our proposed model. We evaluate our model on two public datasets, i.e., Switchboard dialogue act corpus and the MapTask corpus. Extensive experiments show that our proposed model is not only robust, but also achieves better performance when compared with some state-of-the-art baselines. 
	\end{abstract}
	
	
	\section{Introduction}
	Dialogue Act (DA) which represents the meaning of utterances has been widely adopted in computational linguists, especially in the dialogue system. The automatic recognition of DAs is an important step toward understanding spontaneous dialogue, which will facilitate many applications such as human-computer dialogue systems \cite{traum1999speech}, language understanding applications \cite{ezen2015understanding}, spoken language translation \cite{reithinger1996predicting}, or automatic speech recognition \cite{stolcke2000dialogue}. Table \ref{table:dataset} shows a snippet of dialog with utterances and their corresponding labels. In dialogs, each utterance is assigned a unique DA label, drawn from a well-defined set. Thus, DAs can be thought of as a tag set that classifies utterances according to a combination of pragmatic, semantic, and syntactic criteria.
	From Table \ref{table:dataset}, we can also find that knowing the past utterances of dialog can help easing the prediction of the current DA state, thus help to narrow the range of utterance generation topics for the current turn \cite{chen2018dialogue}. For instance, the ``Greeting" and ``Farewell" acts are often followed with another same type utterances, the ``Answer" act often responds to the former ``Question" type utterance. 
	\begin{table}[!t]
		\centering
		\begin{tabular}{lll}
			\hline
			\textbf{Speaker}&\textbf{Utterance}&\textbf{DA label}\\
			\hline
			A & Hi, long time no see. & Greeting \\
			B & Hi, how are you? & Greeting \\
			A & What are you doing these days? &Question \\
			B & I'm busying writing my paper. &Answer \\
			A & I heard that the deadline is coming. &Statement \\
			B & Yeah. &Backchannel \\
			A & You need to make a push.&Opinion \\
			B & Sure, that’s why I am so busy now. &Agreement \\
			A & I can't bother you for too long, goodbye. &Farewell \\
			B &See you later.&Farewell \\
			\hline
		\end{tabular}
		\caption{A snippet of a conversation sample. Each utterance has related dialogue act label \cite{chen2018dialogue}.}
		\label{table:dataset}
	\end{table}
	
	\noindent\textbf{Motivation.} Currently, there have been many research works focusing on the problem of DA classification, ranging from multi-class classification to structured prediction \cite{grau2004dialogue, stolcke2000dialogue}. In many previous works, hand-crafted features are created and fed into a multi-class classifier such as SVM and Naive Bayes \cite{grau2004dialogue,stolcke2000dialogue}, making the model labor intensive and can not be scaled up well across different datasets. 
	To better learn the representation of utterance, recent studies \cite{khanpour2016dialogue,ji2016latent} have applied deep learning based models for the DA recognition task, and have shown promising results.
	However these previous works mainly suffer the limited memory due to the dynamic characteristic of dialog. In other words, each response in dialog is correlated with the previous utterances. 
	For example, it is evident that during a conversation, the speaker's intent is influenced by the former utterance such as the previous ``Greeting" and ``Farewell" examples. This limitation makes these models can't represent dialogues with large numbers of turns and long utterances.
	Another limitation of these previous works lies in the training process of deep neural networks. It has been shown that traditional neural models are often vulnerable to \textit{adversarial examples}, which are examples created by making small perturbations to the input \cite{miyato2016adversarial}.
	
	Motivated by these above mentioned issues, in this paper, we propose a unified framework integrating dynamic memory network and adversarial learning for DA classification. To better represent the state of utterances, we draw some insights from Dynamic Memory Network (DMN) which have been successfully applied in question and answering \cite{xiong2016dynamic}. Unlike previous attention-based deep neural networks, DMN can computes dynamic sentence representations dependently and hence can be easily be used for representation of dialog.
	Specifically, we first formulate the DA classification task into a questioning and answering setting (\textit{Q: What is the label of this utterance?}). Then we propose an improved DMN to learn and memorize the utterance, which contains a hierarchical pyramid utterance encoder. 
	Moreover, we train our model via adversarial training which is a process of training a model to correctly classify both unmodified examples and adversarial examples. It improves not only robustness to adversarial examples, but also generalization performance for original examples.
	The main contributions of this paper can be summarized as follows:
	
	\begin{itemize}
		\item Unlike previous studies, to the best of knowledge, it's the first time that we model the DA classification problem from the perspective of question and answering, and propose an improved dynamic memory network with hierarchical pyramid utterance encoder for better representation of utterances. 
		\item To increase the robustness and generalization of our model, we make small perturbations to the input data, and apply adversarial training to train our proposed model.
		\item Extensive experiments and analysis on two real-world datasets verify the effectiveness of our proposed model when compared with some state-of-the art baselines. 
	\end{itemize}
	
	\noindent\textbf{Organization.} The rest of this paper is organized as follows. In Section \ref{sec_relatedwork}, we provide a brief review of the related work about dialogue act recognition problem. In Section \ref{sec_preliminaries}, we first formulate the problem of dialogue act classification from the viewpoint of question and answering, and introduce some background knowledge about dynamic memory network.
	We elaborate our proposed improved dynamic memory network for DA in Section \ref{sec_model}.
	Extensive experimental results and analysis are presented in Section \ref{sec_experiments}. Finally, we provide some concluding remarks in Section \ref{sec_conclusion}.
	
	\begin{figure*}[!t]
		\centering
		\includegraphics[width=0.98\textwidth]{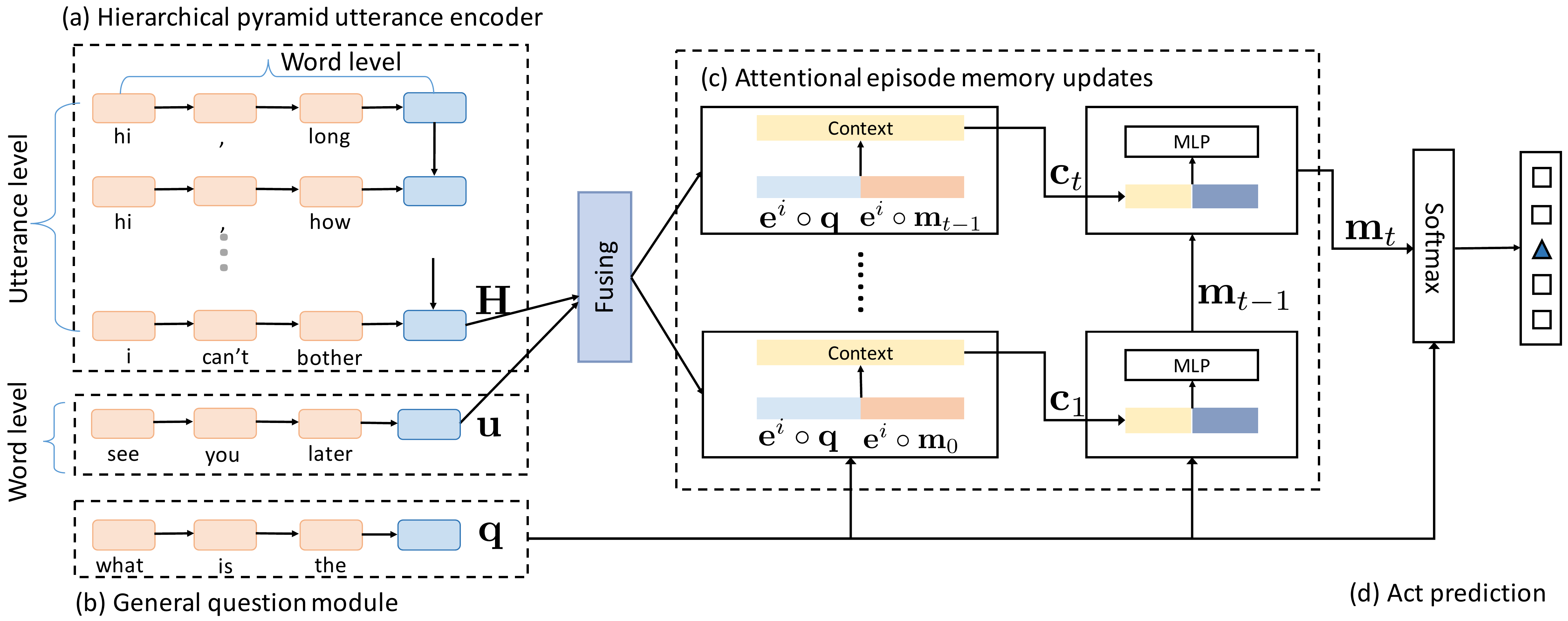}\\
		\caption{An overview of the network architecture of our proposed model. (a) Hierarchical pyramidal utterance encoder module. (b) General question module. (c) Attentional episode memory updates module. (d) Act prediction module.}
		\label{fig:architecture}
	\end{figure*}
	\section{Related Work}\label{sec_relatedwork}
	In this section, we briefly review some related works on dialogue act classification, dynamic memory network and adversarial learning.
	
	\subsection{DA Classification}
	Most of the existing work for the problem of DA classification can be categorized as following two classes: a) Regarding the DA classification as a multi-classification problem \cite{reithinger1997dialogue,geertzen2007multidimensional}. 
	b) Regarding the DA classification as a sequence labeling problem \cite{surendran2006dialog}.
	Recently, approaches based on deep learning methods improve many state-of-the-art techniques in NLP including DA classification accuracy on open-domain conversations \cite{kalchbrenner2013recurrent,khanpour2016dialogue,ji2016latent,lee2016sequential}. Kalchbrenner et al. \cite{kalchbrenner2013recurrent} use a mixture of CNN and RNN to represent utterances where CNNs are used to extract local features from each utterance and RNNs are used to create a general view of the whole dialogue. Khanpour et al. \cite{khanpour2016dialogue} design a deep neural network model that benefits from pre-trained word embeddings combined with a variation of the RNN structure for the DA classification task. Ji et al. \cite{ji2016latent} also investigate the performance of using standard RNN and CNN on DA classification and get the cutting edge results on the MRDA corpus using CNN. Lee et al. \cite{lee2016sequential} propose a model based on CNNs and RNNs that incorporates preceding short texts as context to classify current DAs. Unlike previous models, we cast the DA classification task into a question and answering problem.
	
	\subsection{Dynamic Memory Network} 
	One line of research related to dynamic memory network is attention and memory mechanism \cite{graves2014neural,bahdanau2014neural}, which have been successfully applied in many tasks such as text generation \cite{you2016image, wan2018improving} and question answering \cite{zhao2017video,pan2017memen}. In these works,  memory is encoded as a continuous representation and operations on memory (e.g. reading and writing) are typically implemented with neural networks. Attention mechanism could be viewed as a compositional function, where lower level representations are regarded as the memory, and the function is to assign a weight to each lower position when computing an upper level representation. Such attention based approaches have achieved promising performances on a variety of NLP tasks \cite{luong2015effective}. Based on these works, \cite{kumar2016ask} develops dynamic memory network which simultaneously contains memory updating mechanism and attention mechanism. \cite{xiong2016dynamic} proposes an improved dynamic memory network with some modifications in memory and input module. The dynamic memory network has been successfully applied in many scenarios such as question answering and sentiment analysis. To the best of our knowledge, it's the first time that we apply dynamic memory network in DA classification.  
	
	\subsection{Adversarial Learning}
	Adversarial training \cite{goodfellow2015explaining} introduces an end-to-end and deterministic way of data perturbation by utilizing the gradient information. It could design adversarial examples to attack \cite{tramer2017ensemble} or improve robustness \cite{ wang2018not, zhang2018layerwise} of neural network models. Adversarial training is originally used in the context of image classiﬁcation tasks where the input data is continuous. Miyato et al. \cite{miyato2016adversarial} adopt adversarial training to text classiﬁcation by adding perturbations on word embeddings and also extends it to a semi-supervised setting by minimizing the entropy of the predicted label distributions on unlabeled data. Wu et al. \cite{wu2017adversarial} apply adversarial training in relation extraction within the multi-instance multi-label learning framework. There are also other works on regularizing classifiers by adding random noise to the data, such as dropout \cite{srivastava2014dropout} and its variant for NLP tasks such as word dropout \cite{iyyer2015deep}. In \cite{xie2017data}, Xie et al. discuss various data noising techniques for language models. Sogaard and Li et al. \cite{li2017robust} focus on linguistic adversaries. Inspired by these works, in this paper we apply adversarial training to train our dynamic memory network for the task of DA classification. 
	
	\section{Preliminaries}\label{sec_preliminaries}
	In this section, we first declare some notifications through this paper and formulate the DA classification problem mathematically. We also present some background knowledge on dynamic memory network.
	
	\subsection{Task Description}
	Assume that we have a set $\mathcal{D}$ of $N$ conversations, i.e. $\mathcal{D}=\{C^1, C^2, \ldots, C^N\}$ with $(Y^1, Y^2, \ldots, Y^N)$ corresponding target DAs. For each conversation $C^i$, it consists of a series of utterances, i.e. $C^i=\{u_1,u_2,\cdots,u_d\}$ and the corresponding labels of $C^i$ are $Y^i = \{y_1,y_2,\cdots,y_d\}$, where $d$ is the number of utterances. In other words, for each utterance $u_j$ in each conversation, we have an associated target label $y_j \in \mathcal{Y}$, where $\mathcal{Y}$ is the set of all possible DAs. Each utterance $u_j$ in turn is itself a sequence of $|u_j|$ words stringed together, i.e., $u_j = (w_1, w_2, \ldots w_{|u_j|})$.
	
	In this paper, we cast the DA classification problem into a question and answering setting. Specifically, we first fix the question as \textit{``what is the label of this utterance?"}. Then for a given utterance, our goal is to generate the answer/label for it. For a given conversation $C^i=[u_1, \ldots, u_d]$, $u_j = (w_1, w_2, \ldots w_{|u_j|})$ represents an utterance and $w_n$ denotes the $n$-th word in $u_j$. Let $H=[u_1, \ldots, u_{k-1}]$ denote a dialogue context, the $k-1$ historical utterances, and $u_k$ be a response which means the next utterance. Our goal is to predict the DA label of $u_k$, given its dialogue context $H$.
	
	\subsection{Dynamic Memory Networks}
	Dynamic Memory Network (DMN) is a new neural network architecture based on attention mechanism with the ability of memorizing and reasoning. It has been widely used in question answering tasks since it was first proposed in \cite{kumar2016ask}. In this subsection, we introduce some background knowledge on DMN as presented in \cite{kumar2016ask}. The DMN consists of four modules: input module, question module, episode memory module and answer module. 
	
	\subsubsection{Input Module}
	In this module, the input data corresponding to the question being asked to are encoded into a sequence of distributed vectors representations. We name the vectors as facts (evidences), denoted as $\mathbf{E}=[\mathbf{e}_1, \mathbf{e}_2, \cdots, \mathbf{e}_{|E|}]$, where $|E|$ is the total number of facts, usually is the number of sentence in a document. So $\mathbf{e}_i$ is the vector representation of the $i$-th sentence. The order of vectors in $E$ cannot be changed at random because memory updates need to be based on the order of the facts. Long Short-Term memory (LSTM) \cite{Hochreiter:1997:LSM:1246443.1246450} and Gated Recurrent neural Units (GRU) \cite{chung2014empirical} are typically used to encode the input data. In \cite{kumar2016ask}, GRU is selected as an encoder for the trade-off between computational efficiency and performance effectiveness. 
	The detailed operation of GRU is defined as follows:
	\begin{eqnarray}
	\mathbf{r}_t &=& \sigma(\mathbf{W}_{r}\mathbf{x}_t + \mathbf{W}_{r}\mathbf{g}_{t-1} + \mathbf{b}_r), \nonumber \\
	\mathbf{z}_t &=& \sigma(\mathbf{W}_{z}\mathbf{x}_t + \mathbf{W}_{z}\mathbf{g}_{t-1} + \mathbf{b}_z), \nonumber \\
	\hat{\mathbf{g}}_t &=& \tanh(\mathbf{W}_{g}\mathbf{x}_t + \mathbf{W}_{g}(\mathbf{r}_t\circ \mathbf{g}_t) + \mathbf{b}_g), \nonumber\\ 
	\mathbf{g}_t &=& \mathbf{z}_t \circ \hat{\mathbf{g}}_t + (1 - \mathbf{z}_t) \circ \mathbf{g}_{t-1},
	\end{eqnarray}
	where $\mathbf{r}_t$ and $\mathbf{z}_t$ are the reset gate vector and update gate vector respectively, $\mathbf{g}_t$ is the output vector, $\mathbf{W}$s and $\mathbf{b}$s are weights matrices and biases, $\sigma$ is the sigmoid activation function, $\circ$ is an element-wise multiplication. 
	\subsubsection{Question Module}
	This module is similar to the input module, which also utilizes GRU as the encoder and encodes a sequence of question words into a distributed vector representation $\mathbf{q}$. And then $\mathbf{q}$ is fed into episode memory module and answer module separately. 
	\subsubsection{Episode Memory Module}
	Episode memory module is the main component of the DMN. It is comprised of an attention mechanism and a memory updating mechanism. This module iterates over the input evidences representations and extracts the information to answer the question $\mathbf{q}$. When the questions are too complex to answer so that we need reasoning, the episode memory network may iterates over the input evidences representations multiple times. The episode memory may be updated after each iteration. We denote the memory after $i$-th iteration over the input evidences representations as $\mathbf{m}_i$. We initialize the foremost $\mathbf{m}_0$ as the question vector $\mathbf{q}$. The attention mechanism is in charge of producing a contextual vector $\mathbf{c}_i$, a weighted sum of the input evidences representation, with relevance of the question $\mathbf{q}$ and the previous memory $\mathbf{m}_{i-1}$. The weights of evidences that contribute more to the answer are larger. The episode module may focus on the important information via soft alignment. The memory updating mechanism is in charge of producing the $\mathbf{m}_i$ based on the contextual vector $\mathbf{c}_i$ and the previous memory $\mathbf{m}_{i-1}$. 
	\subsubsection{Answer Module}
	The answer module generates an appropriate answer given the question representation $\mathbf{q}$ and the final episode memory. After multiple updates, $\mathbf{m}_{T}$ contains all information that is required to answer the question. This module takes different execution modes according to different types of tasks. For single-token-answer case, it is treated as a simple classification task. The token-generation layer is composed of a linear layer with a softmax activation classifier to compute the probability distribution of the answer over the entire vocabulary table. For task that requires generating a sequence of tokens, another GRU is used to decode the concatenation of $\mathbf{q}$ and $\mathbf{m}_T$ to a sequence of tokens.
	
	\subsection{An Overview}
	Figure \ref{fig:architecture} shows an overview of the network architecture of our proposed model. The framework can be divided into four parts: (a) Hierarchical pyramid utterance encoder module (cf. Sec. \ref{subsec_pyramid}). In this module, we first represent the input utterance via a pyramidal BiGRU. We first embed each word token into a continuous distributed embedding using \textit{Glove} \cite{pennington2014glove}, and then we add a perturbation for each word embedding for adversarial training. Then we fed the perturbed embedding to a pyramidal BiGRU. For dialogue history embedding, we apply a pyramidal BiGRU to embed each utterance and apply another utterance level GRU to represent the whole context. We call this encoder as hierarchical pyramidal utterance encoder.
	(b) General question module (cf. Sec. \ref{subsec_question}). In this module, we fix the question as \textit{``What is the label of this utterance?"}. We adopt a vanilla BiGRU to embed the question.
	(c) Attentional episode memory updates module (cf. Sec. \ref{subsec_memory}). This module is the core component of our proposed framework which is comprised of attention and memory mechanism. The attention mechanism is to associate a weight to each hidden states (also called facts in DMN). The generated context vector contains the information of facts, question and memory. Then the context vector is fed into the memory updates module for memory updating.
	(d) Act prediction module (cf. Sec. \ref{subsec_output}). This module is the output module of our proposed framework. In our scenario, we use the softmax layer to map the hidden states to the DA label space. We will elaborate each component of this framework in the following sections.

	\section{DMN with Adversarial Training} \label{sec_model}
	In this section, we describe our proposed \underline{D}ynamic \underline{M}emory \underline{N}etwork with \underline{A}dversarial \underline{L}earning (ALDMN) for dialogue act classification. We first describe our dynamic memory network with hierarchical pyramidal utterance encoder and general question module. Then, we describe adversarial training to train our model in detail.
	\subsection{Hierarchical Pyramidal Utterance Encoder} \label{subsec_pyramid}
	\begin{figure}[!t]
		\centering
		\includegraphics[width=0.49\textwidth]{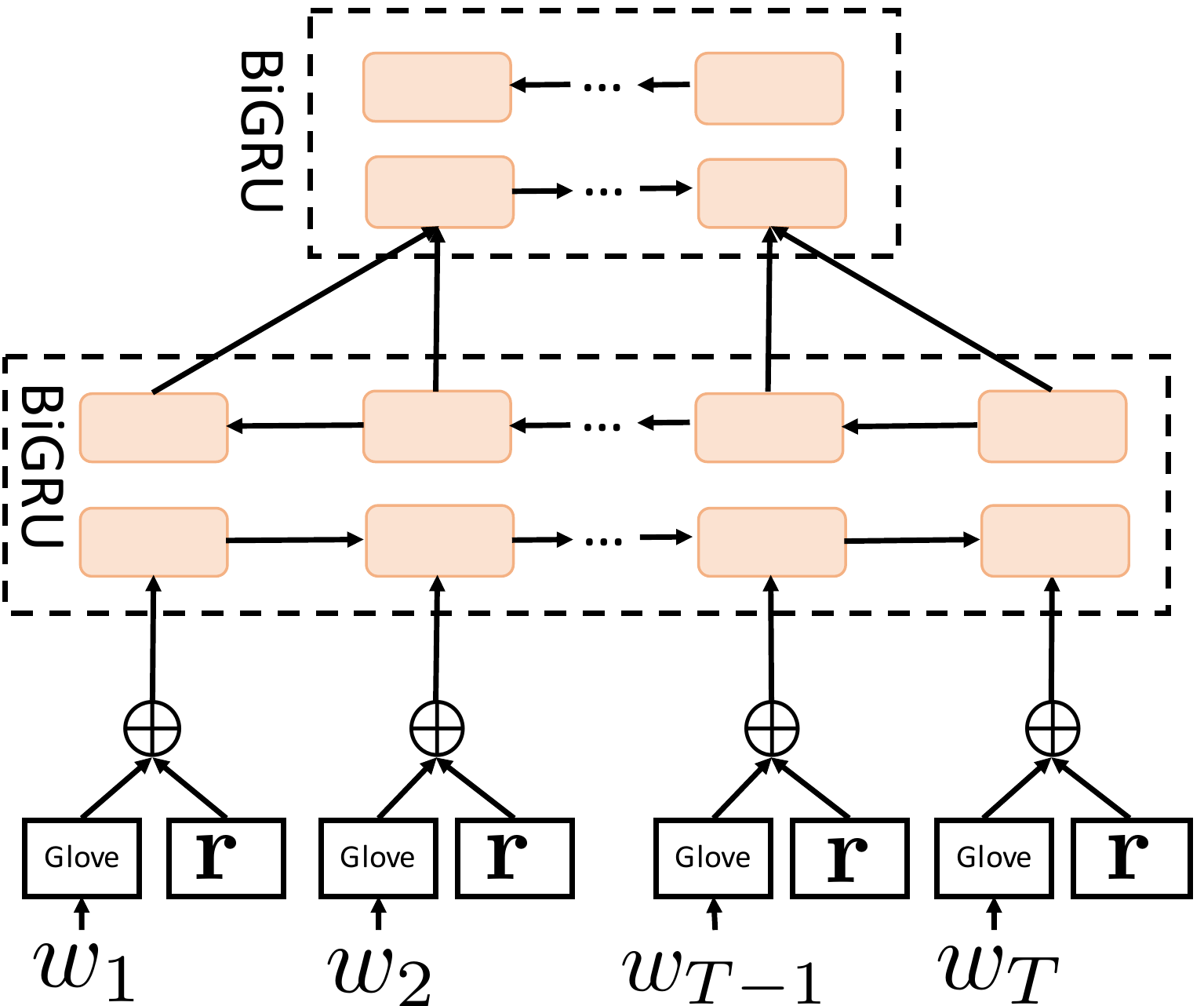}
		\caption{Network structure of pyramidal utterance encoder.}
		\label{fig:sns_heatmap_eg1}
	\end{figure}
	
	In DA classification task, the input is an utterance of sequence of $n$ words $\{w_1,w_2,\ldots,w_n\}$ with its history utterances $H=[u_1,\ldots,u_{k-1}]$. Therefore, the facts of input are composed of two parts (i.e. input utterance facts $\mathbf{e}_u$ and history utterance facts $\mathbf{e}_H$). We model these two parts one by one. 
	
	To encode each utterance in the dialog, there have been many options such as LSTM \cite{kumar2017dialogue}, Gated Recurrent Unit (GRU) \cite{chung2014empirical, xiong2016dynamic}. 
	Pyramidal bidirectional GRU which is an alternative bidirectional GRU that reduces the time dimension after each layer has been successfully applied in character-level sentence encoding and decoding \cite{chan2016listen}. In this paper we try some available encoders, including LSTM, GRU, pyramidal GRU and their bidirectional versions. Experimental results show that the pyramidal bidirectional GRU achieves the best performance, so we adopt it for utterance encoding.
	
	To start with, for the following adversarial training, since the utterance tokens are discrete, we define the perturbation on continuous word embeddings instead of discrete word inputs. Thus, we first use Glove \cite{pennington2014glove} to transform an utterance of $n$ words $\{w_1,w_2,\ldots,w_n\}$ into distributed continuous embeddings, and then add perturbations to them. The word vector of $w_i$ can be represented as follows:
	\begin{equation}
	\mathbf{v}(w_i) = Glove(w_i) + \mathbf{r}_{adv}.
	\end{equation}
	
	We then feed the word embeddings into a pyramidal bidirectional GRU. The hidden states in each bidirectional GRU are defined as follows:
	
	\begin{align}
	\overrightarrow{\mathbf{f}}_t^i &= \text{GRU}(\overrightarrow{\mathbf{f}}_{t-1}^i, \mathbf{e}_t^{i-1}), \\
	\overleftarrow{\mathbf{f}}_t^i  &= \text{GRU}(\overleftarrow{\mathbf{f}}_{t+1}^i, \mathbf{e}_t^{i-1}), \\
	\mathbf{h}_t^i  &= \overrightarrow{\mathbf{f}}_t^i +\overleftarrow{\mathbf{f}}_{t+1}^i ,
	\end{align}
	where GRU denotes the gated recurrent unit function, which has shown to improve the performance of RNNs \cite{cho2014learning,hochreiter1997long}.
	
	\noindent\textbf{Input utterance facts. }For pyramidal Bi-GRU, the input from the previous layer input $\mathbf{e}_t^0 = \mathbf{v}_t^0$ and
	\begin{equation}
	\mathbf{e}_{u,t}^i = \tanh(\mathbf{W}_{pyr}^{(i)}[\mathbf{h}_{2t}^{i-1}; \mathbf{h}_{2t+1}^{i-1}]) + \mathbf{b}_{pyr}^{(i)},
	\end{equation}
	for $i >0$. The weight matrix $\mathbf{W}_{pyr}^{(i)}$ thus reduces the number of hidden states for each additional hidden layer by half, and hence the encoder has a pyramidal structure. At the final hidden layer we obtain the encoded representation $\mathbf{e}$ consisting of $\left \lceil T/2^{N-1} \right \rceil$ hidden states, where $N$ denotes the number of hidden layers. We can find that after the input sequence passes through the pyramidal encoder, the number of encoder output becomes less, which reduces the computational load for the following attention mechanism. 
	
	\noindent\textbf{Historical utterance facts. }After encoding each utterance via pyramidal  Bi-GRU, we apply another GRU to represent the historical utterances:
	\begin{equation}
	\mathbf{e}_{H,t}^{i}= \text{GRU}(\mathbf{e}_{H,t-1}^i, \mathbf{e}_{u,t-1}^{i-1}).
	\end{equation}
	
	Finally, we can denote the input facts by $\mathbf{e}=\{\mathbf{e}_{u}^1, \ldots, \mathbf{e}_{u}^{|u|}, \ldots,\mathbf{e}_{H}^1, \ldots, \mathbf{e}_{H}^{|H|}\}$.
	
	\subsection{General Question Module}\label{subsec_question}
	Different from the question answering task, in DA classification task, none of the utterance is provided with a question. Actually, we can consider all of the utterances facing the same question, such as \emph{``What is the label of this utterance?''}. Thus, we compute a general vector $\mathbf{q}$ representing the same question, where $\mathbf{q}$ is jointly learned with other model parameters and used in the subsequent module.
	
	\subsection{Attentional Episode Memory Updates}\label{subsec_memory}
	Attention mechanism and memory updating mechanism constitute the main components of this module. The episodic memory module, as depicted in Figure \ref{fig:architecture} (c), retrieves information from the input facts $\mathbf{e}=\{\mathbf{e}_{u}^1, \ldots, \mathbf{e}_{u}^{|u|}, \ldots,\mathbf{e}_{H}^1, \ldots, \mathbf{e}_{H}^{|H|}\}$ provided to it by focusing attention on a subset of these facts. We implement this attention by associating a single scalar value, the attention gate $\alpha_t^i$, with each fact $\mathbf{e}^i$ during pass $t$.
	
	Here the implementation of attention mechanism is the same as \cite{bahdanau2014neural}. At the $t$-th iteration, we concatenate the output $\mathbf{e}$ of the pyramidal encoder with previous iteration episodic memory $\mathbf{m}_{t-1}$  and question vector $\mathbf{q}$, and then employ the basic soft attention to obtain the $t$-th contextual vector as:
	\begin{align}
	\mathbf{z}_t^i &=[\mathbf{e}^i \circ \mathbf{q}; \mathbf{e}^i \circ \mathbf{m}_{t-1}],\\
	\mathbf{\beta}_t^i  &= \mathbf{W}^{(2)}\tanh \left(\mathbf{W}^{(1)}\mathbf{z}_t^i+\mathbf{b}^{(1)} \right)+\mathbf{b}^{(2)},\\
	\mathbf{\alpha}_t^i &=\frac{\exp(\mathbf{\beta}_t^i)}{\sum_{j=1}^n\exp(\mathbf{\beta}_t^j)}, \\
	\mathbf{c}_t &=\sum_{i=1}^n\alpha_t^i \mathbf{e}^i,
	\end{align}
	where $\circ$ is the element-wise multiplication, $[\cdot;\cdot]$ is the concatenation operation, $\mathbf{W}$ and $\mathbf{b}$ are model parameters. In \cite{xiong2016dynamic}, $\mathbf{z}_t^i$ is set to $\mathbf{z}_t^i=[\mathbf{e}^i \circ \mathbf{q}; \mathbf{e}^i \circ \mathbf{m}_{t-1}; |\mathbf{e}^i - \mathbf{q}|; |\mathbf{e}^i - \mathbf{m}_{t-1}|]$, where $|\cdot|$ is the element-wise absolute value. However in our scenario, $\mathbf{z}_t^i =[\mathbf{e}^i \circ \mathbf{q}; \mathbf{e}^i \circ \mathbf{m}_{t-1}]$ achieves better performance from the experimental results.
	
	Following the memory update component used in \cite{peng2015towards}, we first concatenate the previous episodic memory $\mathbf{m}_{t-1}$, the current contextual vector $\mathbf{c}_t$ and question vector $\mathbf{q}$, and use a ReLU layer for the memory update:
	\begin{equation}
	\mathbf{m}_t=\text{ReLU}(\mathbf{W}^{(3)}[\mathbf{m}_{t-1};\mathbf{c}_t;\mathbf{q}]+\mathbf{\mathbf{b}}^{(3)}), 
	\end{equation}
	where $[\cdot;\cdot]$ is the concatenation operation, $\mathbf{W}^{(3)}$ and $\mathbf{b}^{(3)}$ are model parameters. We use ReLU activation and initialize the memory vector as the question vector: $\mathbf{m}_0=\mathbf{q}$. 
	
	\subsection{Act Prediction}\label{subsec_output}
	The final episodic memory $\mathbf{m}_T$ and question vector $\mathbf{q}$ are concatenated to compute the probability of the correct answer. In DA classification, the answer is the label of the corresponding utterance, which is different from the question answering task predicting a sequence token representing the correct answer. 
	The classifier takes the episodic memory $\mathbf{m}_T$ and question vector $\mathbf{q}$ as input:
	
	\begin{equation}
	p(y|\mathbf{m}_T, \mathbf{q})=\mathrm{softmax}(\mathbf{W}^{(4)}[\mathbf{q};\mathbf{m}_T]+\mathbf{b}^{(4)}),
	\end{equation}
	where $y$ represents the index of multi-class candidate answer. $\mathbf{W}^{(4)}$ and $\mathbf{b}^{(4)}$ are learnable parameters to be optimized in the model.
	
	The loss function is the negative log-likelihood of the true class labels $y$ for each utterance:
	\begin{equation}
	\mathcal{L}(\mathbf{u},\theta) =-\sum_{i=1}^{M}\log\ p(y^{(i)}|\mathbf{u}; \mathbf{\theta}),
	\end{equation}
	where $M$ is the number of utterances in training dataset.
	\subsection{Adversarial Training} \label{adversarial}
	For classification task, many existing models use dropout or $\ell_2$ for model regularization and achieve a good performance. However, these models demonstrate poor performance on adversarial examples even sometimes the perturbations are so small that humans cannot detect it. Adversarial learning is another regularization method of the model, which aims to train the model to correctly classify adversarial examples and real examples. Following the method of adding perturbations in \cite{miyato2016adversarial}, we create continuous perturbations by adding noise to word embedding. Let $\mathbf{u}$ denote the input utterance and $\theta$ the parameters of model. When applied to a model, adversarial training adds the following term to the loss function:
	
	\begin{equation}
	\mathcal{L}_{adv}(\mathbf{\theta}) = -\sum_{i=1}^{M}\log p(y_i|\mathbf{u}_i+\mathbf{r}_{adv,i}; \mathbf{\theta}),
	\end{equation}
	where $M$ is the total number of utterances in training data, $\mathbf{r}_{adv}$ is the worst case perturbations against the current model which can be defined as follows:
	\begin{equation}
	\mathbf{r}_{adv} = \underset{\mathbf{r},||\mathbf{r}||<\epsilon}{\mathrm{argmin}}\, \log p(y|\mathbf{u}+\mathbf{r};\hat{\mathbf{\theta}}) \label{r_adv},
	\end{equation}
	where $\mathbf{r}$ is a perturbation on the input utterance $\mathbf{u}$ and $\hat{\theta}$ is a constant set to the current parameters of a classifier, indicating that the backpropagation algorithm will be used to propagate gradients through the adversarial example construction process.
	
	We can see that the $\mathbf{r}_{adv}$ can not be calculated exactly in general, since exact minimization with respect to $\mathbf{r}$ is intractable for many models such as neural networks. Inspired by \cite{goodfellow2015explaining}, we can approximate this value by linearizing $\log (y|\mathbf{u};\theta)$ around $\mathbf{u}$. Therefore, with a linear approximation and a $\ell_2$ norm constraint in Eq.\ref{r_adv}, the resulting adversarial perturbation is formulated as follows:
	
	\begin{equation}
	\mathbf{r}_{adv} = \frac{\epsilon \mathbf{g}}{||\mathbf{g}||},
	\end{equation}
	where $\mathbf{g}$ is the normalized gradient of log-likelihood with respect to $\mathbf{u}$, which can be formulated as follows:
	\begin{equation}
	\mathbf{g}=\nabla_\mathbf{u}\log\ p(y|\mathbf{u};\hat{\mathbf{\theta}}).
	\end{equation}
	
	We can compute the perturbation by backpropagation in neural networks.  
	To optimize the objective, we employ the Adam \cite{kingma2014adam}, an algorithm for first-order gradient-based optimization of stochastic objective functions, based on adaptive estimates of lower-order moments.
	
	\section{Experiments and Analysis}\label{sec_experiments}
	In this section, we conduct extensive experiments and analysis on two real-world datasets to show the effectiveness of our proposed approach. 
	\subsection{Datasets}  
	We evaluate the performance of our model on two benchmark datasets used in several prior studies for the DA classification task, i.e.  Switchboard Dialogue Act Corpus (SwDA) \cite{godfrey1992switchboard} and MapTask Dialogue Act Corpus (MapTask) \cite{anderson1991hcrc}.  
	
	\textbf{SwDA:} This dataset consists of 1,155 telephone conversations and each of the utterance in the dialogues is mapped into 42 distinguished utterance types via DAMSL taxonomy. We shuffle the data randomly and use 1,050 conversations for training and 105 conversations for testing. 
	
	\textbf{MapTask:} This dataset comprises of $128$ dialogues and more than 27,000 utterances. Each of the utterance is labeled with one of the 13 tags. Unlike SwDA, the MapTask corpus emphasizes on directions and conductions.
	
	\begin{table}[!t]
		\centering
		\begin{tabular}{|l|l|l|l|l|l|}
			\hline
			Dataset &$|C|$&$|V|$&MinL&MeanL&MaxL\\
			\hline
			SwDA&42&19K&16&56&118\\
			MapTask&13&15K&2&4&21\\
			\hline
		\end{tabular}
		\caption{$|C|$ is the number of Dialogue Act classes, $|V|$ is the vocabulary size. MinL, MeanL and MaxL indicate the minimum, mean and maximum of utterance length, respectively.}
		\label{table:stastic}
	\end{table}
	
	\begin{figure}[!t]
		\centering
		\begin{subfigure}[b]{0.49\textwidth}
			\includegraphics[width=\textwidth]{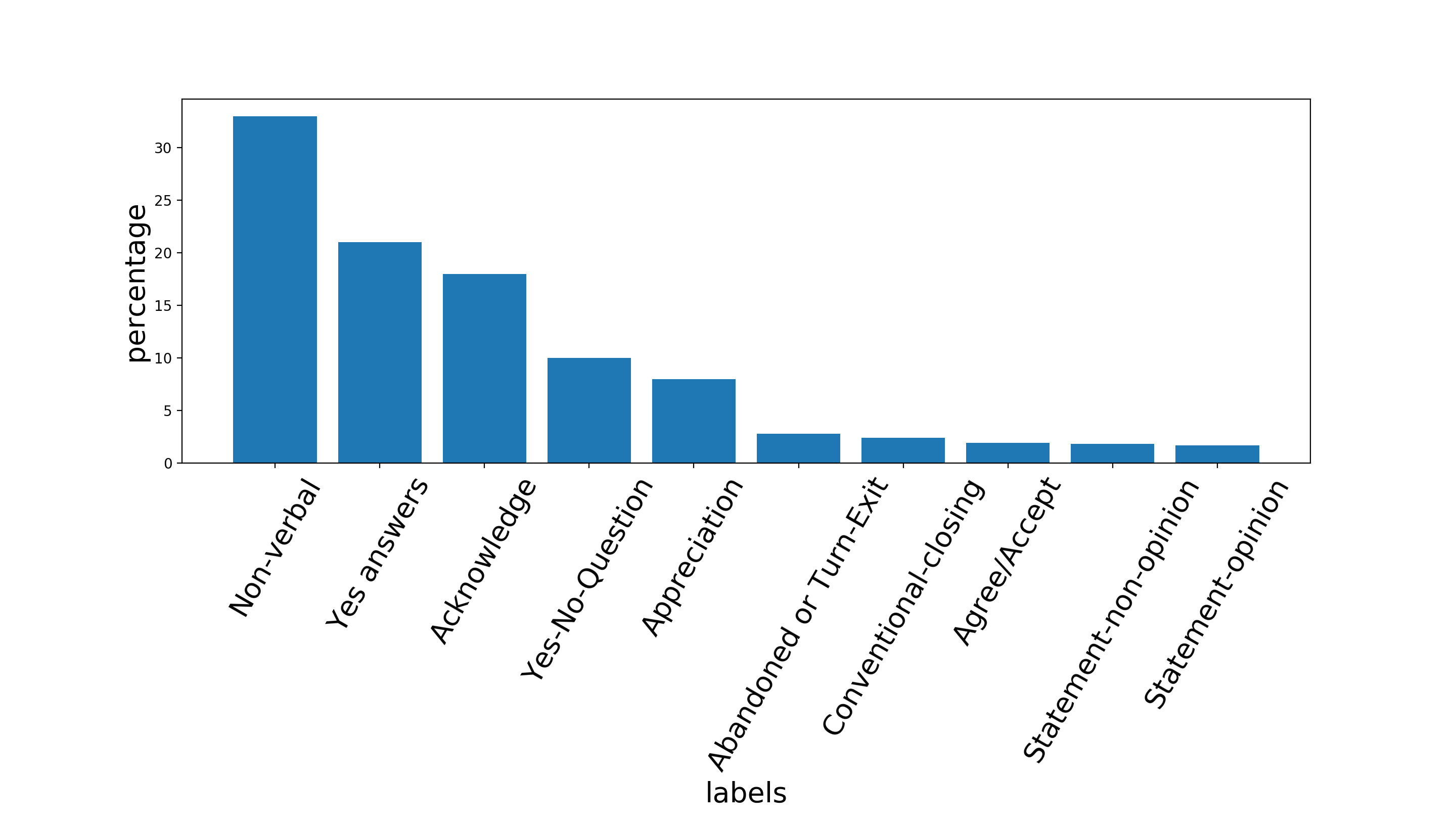}
			\caption{SwDA}
			\label{fig_varadv_swda}
		\end{subfigure}
		\begin{subfigure}[b]{0.49\textwidth}
			\includegraphics[width=\textwidth]{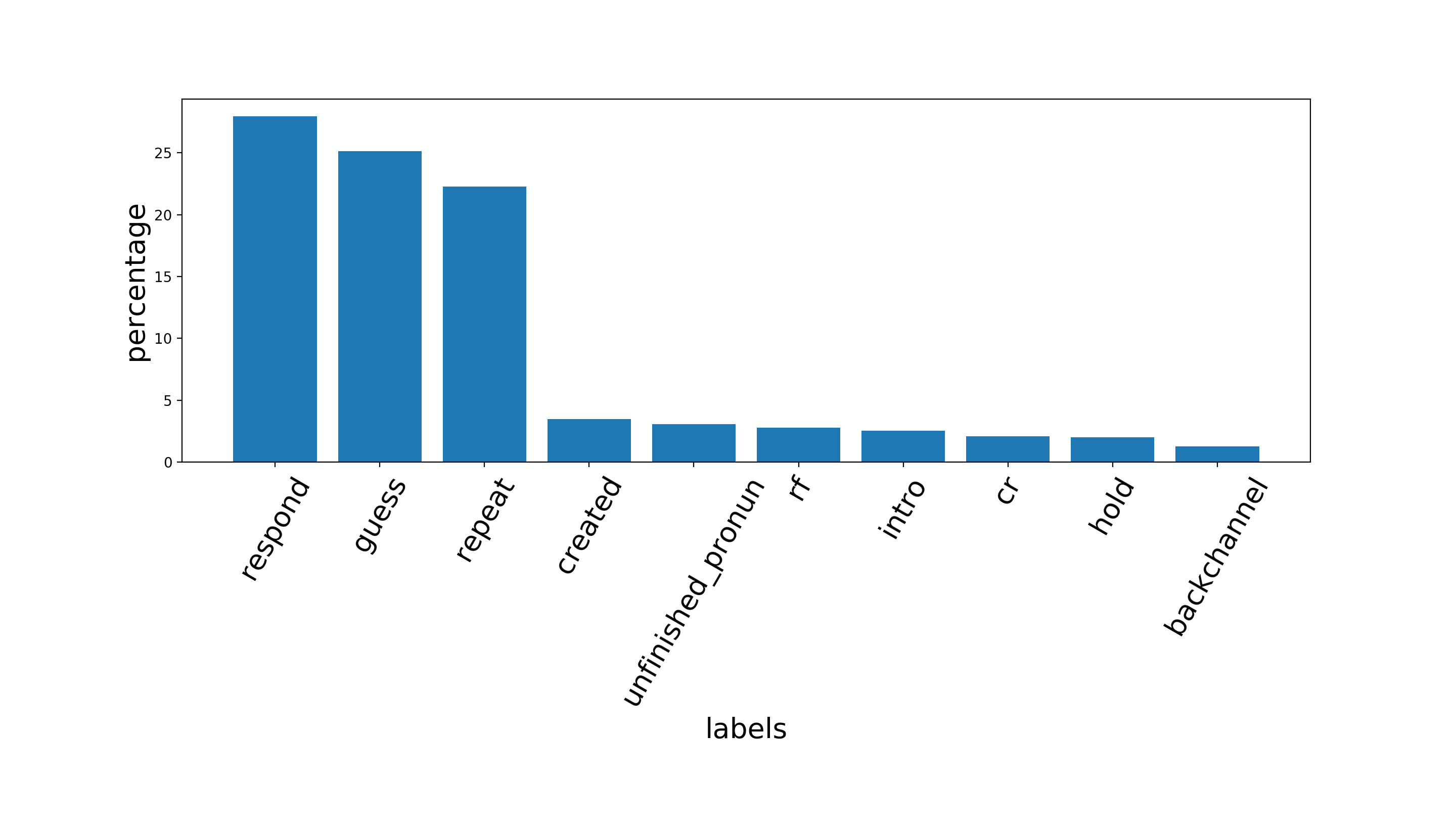}
			\caption{MapTask}
			\label{fig_varadv_maptaskr}
		\end{subfigure}
		\caption{The distribution of utterance labels on two datasets: a) SwDA, b) MapTask.}
		\label{fig_length_distribution}
		\vspace{-0.8em}
	\end{figure}

	Table \ref{table:stastic} presents different statistics for both datasets. For SwDA, training and testing sets are provided but not the validation set, so we use the standard practice of taking a part of training data set as validation set \cite{lee2016sequential}. We also shows the top 10 utterance labels on different datasets in Figure \ref{fig_length_distribution}.
	
	\subsection{Evaluation Metrics}
	We employ the standard accuracy as the metric of evaluating our proposed ALDMN method. The accuracy is defined as: 
	
	\begin{equation}
	Accuracy = \frac{1}{M} \sum_{i=1}^{M}\textbf{1}[\hat{y_i} = y_i],
	\end{equation}
	where $\hat{y_i}$ and $y_i$ are the predicted label and ground true label, respectively. $\textbf{1}[\cdot]$ is the indicator function. When the predicted label is the same as the true, label $\textbf{1}[\cdot]$ equals to $1$ otherwise $0$.
	
	\subsection{Implementation Details}
\begin{figure}[!t]
		\centering
		\includegraphics[width=0.49\textwidth]{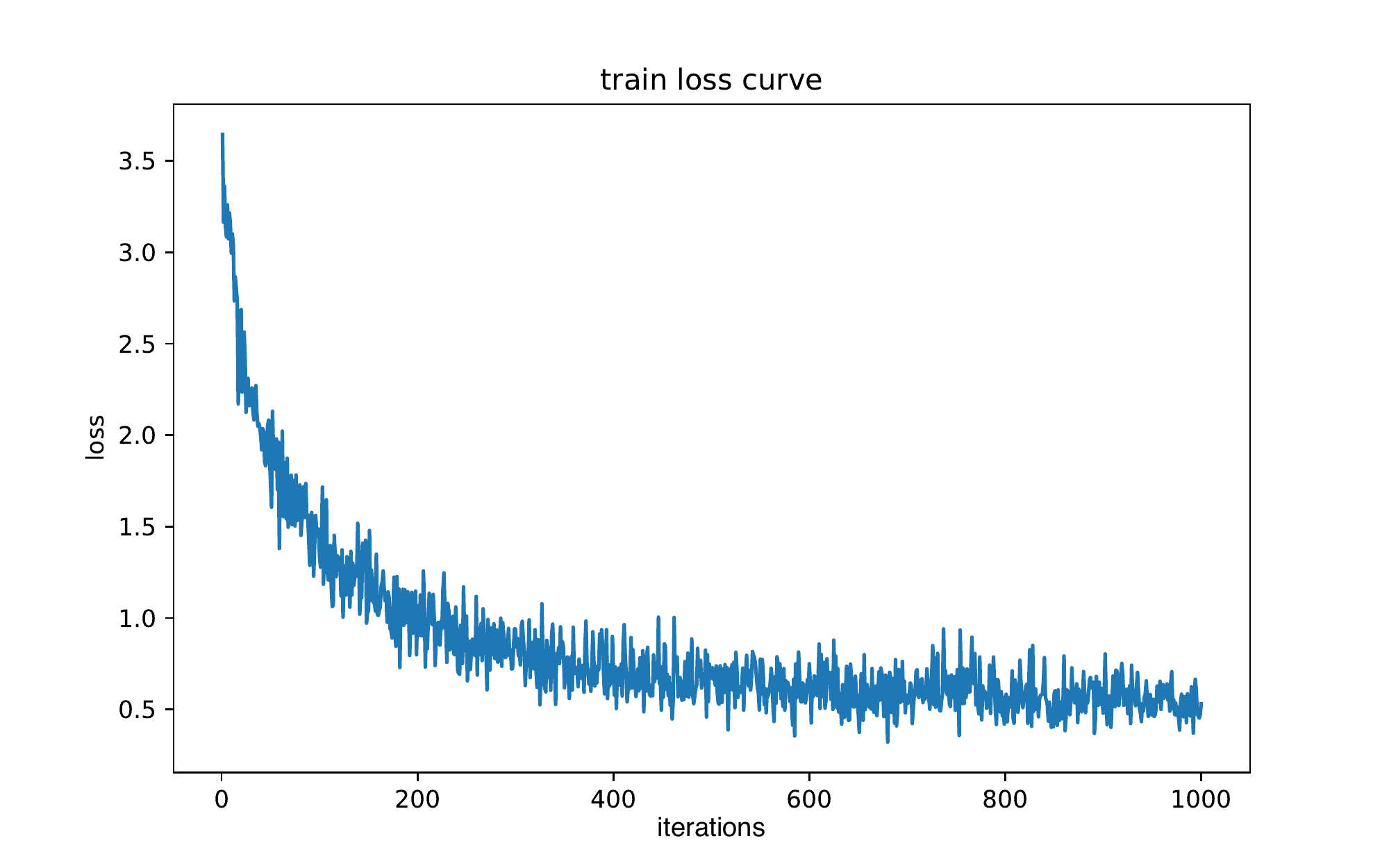}
		\caption{Training loss of each iteration.}
		\label{fig:training_loss}
	\end{figure}

	To train our proposed model, we first randomize the training data and set the mini-batch to $128$. For each batch, the utterance is padded with a special token $<$pad$>$ to the maximum length. We regard the words whose occurrence is less than 2 as Out of Vocabulary (OOV) tokens and replace them with $<$unk$>$s. We remove all punctuation marks except interrogation and all characters are converted to lower-case. We set the memory updating iterations to $3$. We adopt Adam \cite{kingma2014adam} optimizer with the learning rate initialized to $0.01$. We run the training data set for $45$ epochs with early stopping when validation loss did not decline for five consecutive epochs. The random uniform initialization with range $[-0.1, 0.1]$ is used for all matrix variables, including word embedding and other weights. Both the embedding and hidden dimensions are set to $d = 200$. The dropout rate is set to $0.2$. Figure \ref{fig:training_loss} is the training loss curve of our method. We can see that our model is converged after around $300$ iterations.
	
	All the experiments in this paper are implemented with Python 2.7 based on TensorFlow, and run on a computer with an 2.2 GHz Intel Core i7 CPU, 64 GB 1600 MHz DDR3 RAM, and a Titan X GPU with 12 GB memory, running Ubuntu 16.04.
	
	\subsection{ Results and Analysis}
	We compare our proposed method with several other state-of-the-art methods for the problem of dialogue act classification as follows:
	\begin{itemize}
		\item UCI \cite{liu2017using} method embeds contextual information of utterance via hierarchical CNN/RNN for DA classification.
		\item PDI \cite{tran2017aprocessings} method predicts the next label based on the current label probability distribution to avoid label bias.
		\item DRLM-Conditional ~\cite{ji2016a} method is a latent variable recurrent neural network architecture for jointly modeling utterances and DA labels.
		\item BiLSTM-Softmax \cite{khanpour2016dialogue} method uses a bidirectional LSTM to embed utterances and then feeds them to a softmax classifier.
		\item RCNN~\cite{kalchbrenner2013recurrent}. Different from BiLSTM-Softmax, this method uses hierarchical CNN to embed  utterances.
		\item DMN \cite{kumar2016ask} methods is a DMN based method for sentiment classification, where BiLSTM is used for sentences embedding.
		\item ADC \cite{julia2010dialog} and PDI ~\cite{tran2017aprocessings} methds are acoustic and discourse classification based on HMM and SVM.
		\item GAN \cite{tran2017bgenerative} method is a generative neural network which incorporates attention technique and a label-to-label connection.
	\end{itemize}
	Table \ref{table:result_1} and Table \ref{table:result_2} respectively show the experimental \textit{Accuracy} results of the methods on the SwDA and MRDA datasets. The hyper-parameters and parameters which achieve the best performance on the validation set are chosen to conduct the testing evaluation. From these two tables, we can find that our proposed model ALDMN definitely outperforms other baselines in both datasets. On SwDA dataset, our model improves the accuracy over the best performing traditional method UCI by 1.6\%. On MapTask dataset, our model improves the accuracy over the best performing traditional method PDI by 2.6\%. This verifies the effectiveness of our proposed DMN module and adversarial training approach for DA classification.
	\begin{table}[!t]
		\centering
		\begin{tabular}{p{5.4cm}p{1.2cm}}
			\hline
			Model &Accuracy(\%)\\
			\hline
			RCNN(Bulnsom et al. 2013)&73.9\\
			BiLSTM-Softmax(Khanpour et al. 2016)&75.8\\
			DRLM-Conditional(Ji et al. 2016)&77.0\\
			PDI(Tran et al. 2017)&75.6 \\
			UCI(Liu et al. 2017)&79.9\\
			DMN(Kumar et al., 2015) & 75.2\\
			\hline
			ALDMN (Our Model) & \textbf{81.5} \\
			\hline
		\end{tabular}
		\caption{Classification accuracy on SwDA corpus, comparing our ALDMN model with other methods as described in literatures.}
		\label{table:result_1}
	\end{table}
	
	\begin{table}[!t]
		\centering
		\begin{tabular}{p{5.4cm}p{1.2cm}}
			\hline
			Model &Accuracy(\%)\\
			\hline
			ADC(Julia et al. 2010)&55.4\\
			GAN(Tran et al. 2017b)&62.9 \\
			PDI(Tran et al. 2017a)&65.9\\
			DMN(Kumar et al., 2015)&64.7\\
			\hline
			ALDMN (Our Model) & \textbf{68.5} \\
			\hline
		\end{tabular}
		\caption{Classification accuracy on MapTask corpus, comparing our ALDMN model with other methods as described in literatures.}	
		\label{table:result_2}
	\end{table}

	\subsection{Impact of Pyramidal Encoder}
\begin{figure*}[!t]
		\centering
		\begin{subfigure}[b]{3.5in}
			\includegraphics[width=\textwidth]{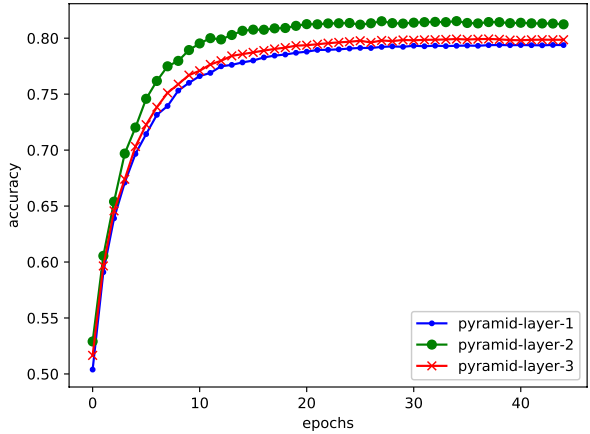}
			\caption{SwDA}
			\label{fig_varpyramidt_swda}
		\end{subfigure}
		\begin{subfigure}[b]{3.5in}
			\includegraphics[width=\textwidth]{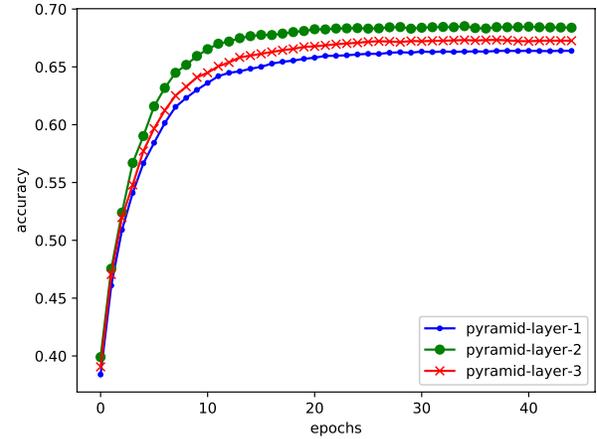}
			\caption{MapTask}
			\label{fig_varpyramidt_maptask}
		\end{subfigure}
		\caption{Classification accuracy curves with different pyramid layer of pyramid-layer-1, pyramid-layer-2, pyramid-layer-3 on the two different datasets: a) SwDA, b) MapTask.}
		\label{fig_varpyramid}
		\vspace{-0.8em}
	\end{figure*}

	To analyze the impact of pyramidal layer, we vary the layers number of pyramidal encoder from 1 to 3. When the layer number is 1, the pyramidal encoder reduced into a BiGRU encoder. Figure \ref{fig_varpyramid} shows the classification accuracy with varying pyramid layers on SwDA and MapTask dataset. 
	From this figure, we can find that our model achieves the best performance when the pyramid layer is set to $2$. This verifies the effectiveness of the pyramidal encoder. Moreover, the performance decrease when the pyramid layer is set to $2$. This can be illustrated by the fact that our encoder may be over-fitting with the number of layers increasing.
	
	\begin{figure*}[!t]
		\centering
		\begin{subfigure}[b]{3.5in}
			\includegraphics[width=\textwidth]{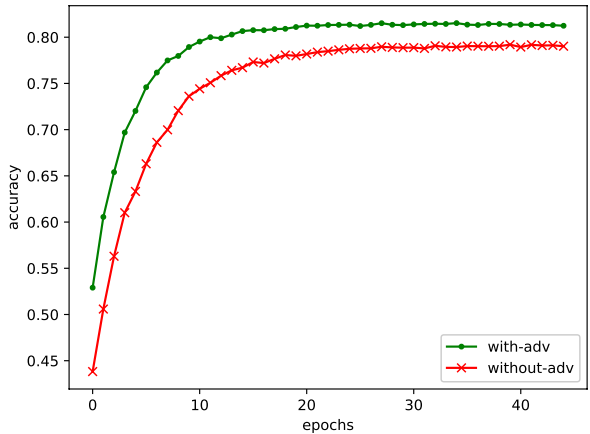}
			\caption{SwDA}
			\label{fig_varadv_swda}
		\end{subfigure}
		\begin{subfigure}[b]{3.5in}
			\includegraphics[width=\textwidth]{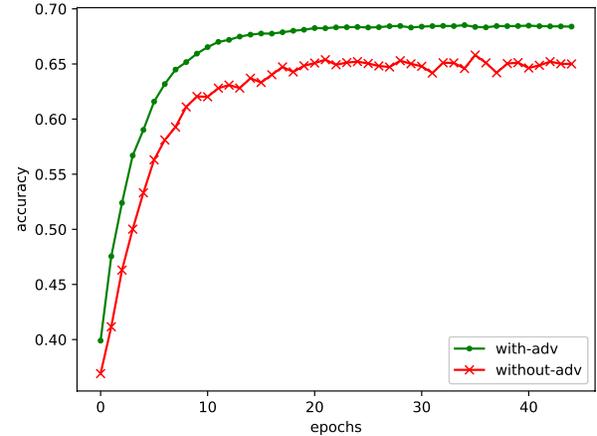}
			\caption{MapTask}
			\label{fig_varadv_maptaskr}
		\end{subfigure}
		\caption{Classification accuracy curves of with (green) and without (red) adversarial learning on two different datasets: a) SwDA, b) MapTask.}
		\label{fig_varadv}
		\vspace{-0.8em}
	\end{figure*}
	\subsection{Impact of Adversarial Learning}
	To demonstrate the effectiveness of adversarial learning on word embeddings, we compare architectures with/without adversarial learning. For adversarial training, we set $\epsilon = 3$. For model without adversarial training, the dropout is applied. Comparing performances are shown in Figure ~\ref{fig_varadv}. From this figure, we can find that our model with adversarial can really improve the accuracy performance in both two datasets. Take SwDA as an example, with only dropout applied, our model achieves accuracy of 79.1\%, while the accuracy of model with adversarial learning can reach up to 81.5\%. This can be explained by the fact that the model without adversarial learning is strongly affected by the syntax of the dataset. In fact, in DA classification task, as the conversations are sometimes recorded in a casual setting, the grammar of utterances are not so strict compared to a formal document. What's more, due to the model pre-training step, some antonyms, although greatly different in semantics, are very close in the vector space. For example, \emph{``big''} is opposite to \emph{``small''} in semantics, but they are sometimes in the similar context. So they are assigned the similar word embeddings on the baseline model while the embeddings cannot convey the actual meaning of these two words. Adversarial learning ensures the representation of sentences not be changed under some perturbations so that those words in same context with different meanings can be separated in the vector space.
	
	\subsection{Performance on Various Data Distribution}
\begin{figure}[!t]
		\centering
		\includegraphics[width=3.6in]{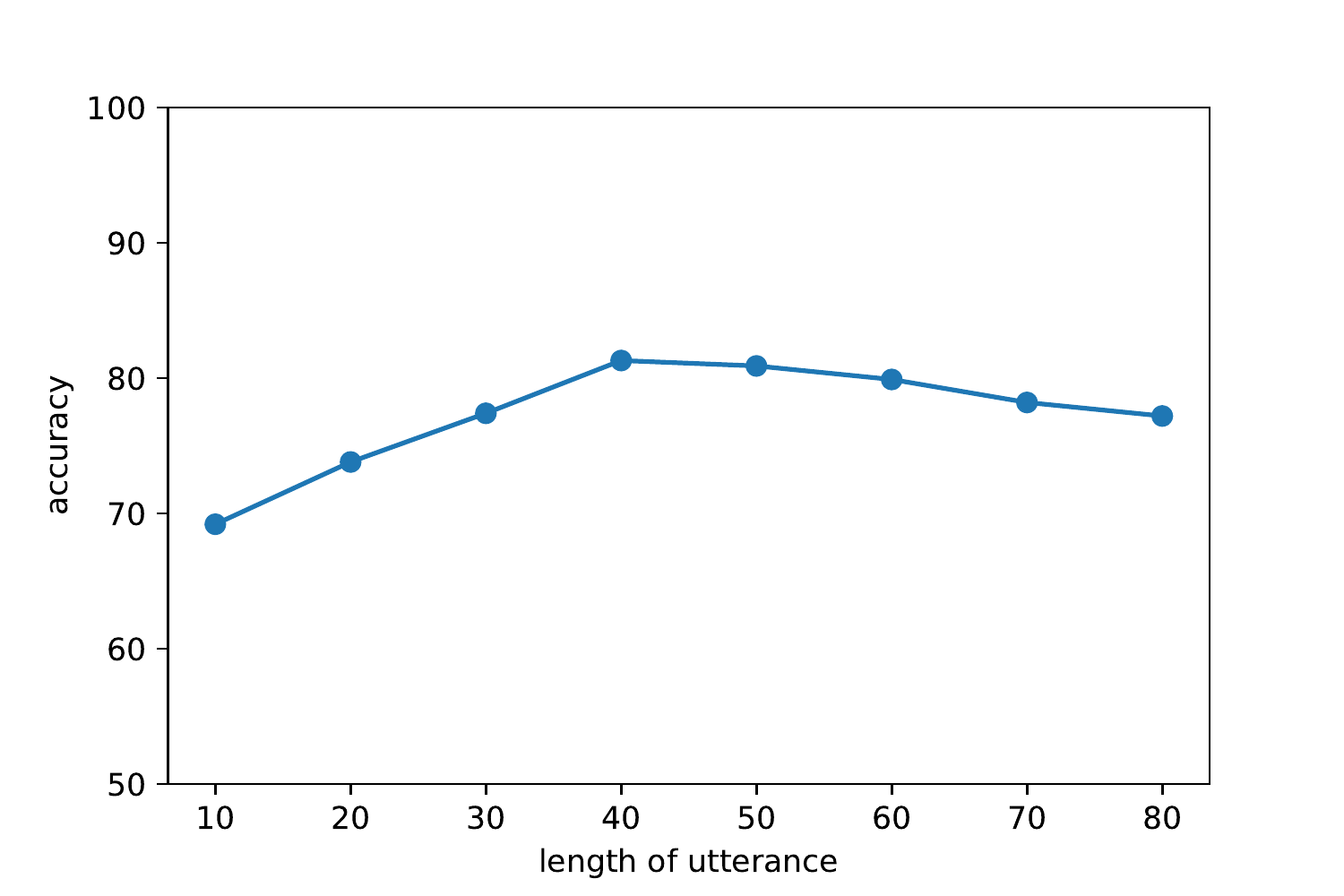}
		\caption{The performance of our proposed model w.r.t. varying length of utterances on SwDA dataset.}
		\label{fig_var_length_performance}
	\end{figure}

	We vary the length of utterances since the utterance length may have an effect on the representation. Figure \ref{fig_var_length_performance} shows the performance of our proposed model w.r.t. varying utterance on SwDA dataset. From this figure, we can find that our model achieves the best performance on the utterances whose length is around $40$. This can be illustrated by the fact that short utterances always contain limited information while long utterances may contain more noises.
	
	\subsection{Confusion Matrix Visualization}
	\begin{figure}[!t]
		\centering
		\includegraphics[width=0.49\textwidth]{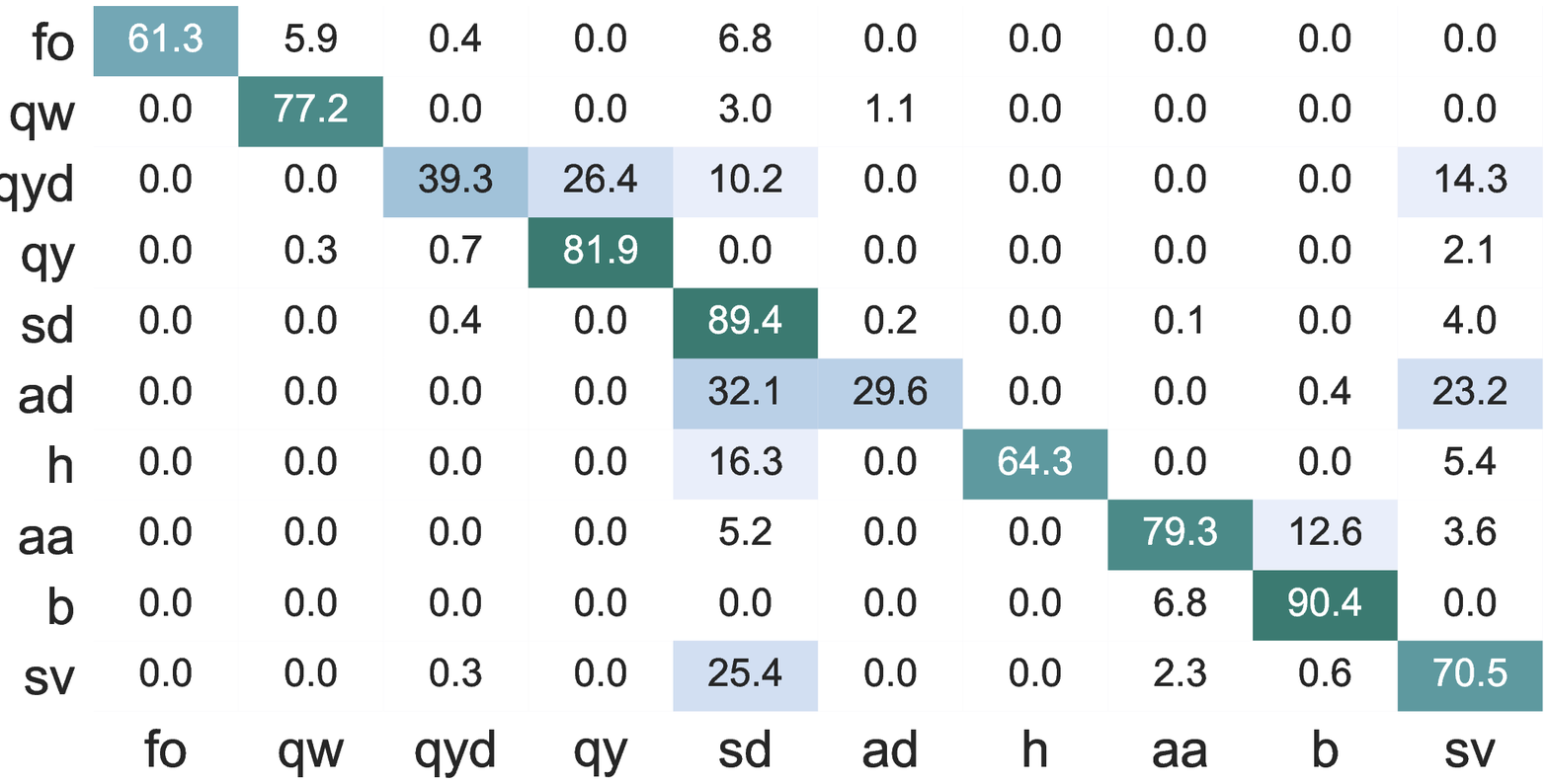}
		\caption{Confusion matrix of our proposed model for the SwDA dataset, where the row denotes the true label and the column denotes the predicted label. The numbers in the bracket besides the DA label in the ﬁrst cell of each row is the count of the number of utterances of that DA label.}
		\label{fig:sns_heatmap_eg1}
	\end{figure}
	Figure \ref{fig:sns_heatmap_eg1} shows the confusion matrix of our proposed model for the SwDA dataset. Among them the most confused pairs are (sd, sv) and (aa, b) which represent (statement-non-opinion, statement-opinion) and (agree-accept, acknowledge) respectively. The total number of utterances with DA `sd', `sv', `aa', and `b' are 72,824, 25,197, 10,820 and 37,096, respectively.  3,423 utterances (4.7\%) with true label non-opinion are predicted incorrectly as opinion, whereas, 64,847 utterances (89.7\%) with true label non-opinion are predicted correctly. Similarly, 6,478 utterances (25.4\%) with true label opinion are predicted incorrectly as non-opinion whereas 17,638 utterances (70.5\%) with true label opinion are predicted correctly. On further analysis of the cause of this confusion between these two class pairs, we identify that some utterances are classified correctly by the model. However, they are marked incorrectly classified because of bias in the ground truth. For some of the utterances, classes are not distinguishable even by humans because of the subjectivity.
	
	\section{Conclusion}\label{sec_conclusion}
	In this paper, we propose a unified framework integrating dynamic memory network and adversarial training for the task of dialogue act classification. Specifically, we first cast the act classification task into a question and answering problem and propose an improved dynamic memory network to represent and memorize the utterance with its corresponding history utterances. In addition, we apply adversarial training to train a more robust model. We demonstrate the effectiveness of our proposed model using the well-known public datasets SwDA and MapTask. Extensive experiments demonstrate that our model can achieve better performance than several state-of-the-art solutions to the problem.
	
	\section{Acknowledgement}
	This work is partially supported by the Ministry of Education of China under grant of No.2017PT18, the Natural Science Foundation of China under grant of No. 61672453, 61773361, 61473273, 61602405, the WE-DOCTOR company under grant of No. 124000-11110 and the Zhejiang University Education Foundation under grant of No. K17-511120-017, No. K17-518051-021. This work is also supported by CCF-Tencent Open Research Fund, NSF through grants IIS-1526499, IIS-1763325, CNS1626432, and NSFC 61672313.
	\bibliographystyle{IEEEtran}
	\bibliography{ref}

\begin{thebibliography}{10}
\providecommand{\url}[1]{#1}
\csname url@samestyle\endcsname
\providecommand{\newblock}{\relax}
\providecommand{\bibinfo}[2]{#2}
\providecommand{\BIBentrySTDinterwordspacing}{\spaceskip=0pt\relax}
\providecommand{\BIBentryALTinterwordstretchfactor}{4}
\providecommand{\BIBentryALTinterwordspacing}{\spaceskip=\fontdimen2\font plus
\BIBentryALTinterwordstretchfactor\fontdimen3\font minus
  \fontdimen4\font\relax}
\providecommand{\BIBforeignlanguage}[2]{{%
\expandafter\ifx\csname l@#1\endcsname\relax
\typeout{** WARNING: IEEEtran.bst: No hyphenation pattern has been}%
\typeout{** loaded for the language `#1'. Using the pattern for}%
\typeout{** the default language instead.}%
\else
\language=\csname l@#1\endcsname
\fi
#2}}
\providecommand{\BIBdecl}{\relax}
\BIBdecl

\bibitem{traum1999speech}
D.~R. Traum, ``Speech acts for dialogue agents,'' in \emph{Foundations of
  rational agency}.\hskip 1em plus 0.5em minus 0.4em\relax Springer, 1999, pp.
  169--201.

\bibitem{ezen2015understanding}
A.~Ezen-Can and K.~E. Boyer, ``Understanding student language: An unsupervised
  dialogue act classification approach,'' \emph{Journal of Educational Data
  Mining (JEDM)}, vol.~7, no.~1, pp. 51--78, 2015.

\bibitem{reithinger1996predicting}
N.~Reithinger, R.~Engel, M.~Kipp, and M.~Klesen, ``Predicting dialogue acts for
  a speech-to-speech translation system,'' in \emph{Spoken Language, 1996.
  ICSLP 96. Proceedings., Fourth International Conference on}, vol.~2.\hskip
  1em plus 0.5em minus 0.4em\relax IEEE, 1996, pp. 654--657.

\bibitem{stolcke2000dialogue}
A.~Stolcke, K.~Ries, N.~Coccaro, E.~Shriberg, R.~Bates, D.~Jurafsky, P.~Taylor,
  R.~Martin, C.~V. Ess-Dykema, and M.~Meteer, ``Dialogue act modeling for
  automatic tagging and recognition of conversational speech,''
  \emph{Computational linguistics}, vol.~26, no.~3, pp. 339--373, 2000.

\bibitem{chen2018dialogue}
Z.~Chen, R.~Yang, Z.~Zhao, D.~Cai, and X.~He, ``Dialogue act recognition via
  crf-attentive structured network,'' in \emph{The 41st International ACM SIGIR
  Conference on Research \& Development in Information Retrieval}.\hskip 1em
  plus 0.5em minus 0.4em\relax ACM, 2018, pp. 225--234.

\bibitem{grau2004dialogue}
S.~Grau, E.~Sanchis, M.~J. Castro, and D.~Vilar, ``Dialogue act classification
  using a bayesian approach,'' in \emph{9th Conference Speech and Computer},
  2004.

\bibitem{khanpour2016dialogue}
H.~Khanpour, N.~Guntakandla, and R.~Nielsen, ``Dialogue act classification in
  domain-independent conversations using a deep recurrent neural network,'' in
  \emph{Proceedings of COLING 2016, the 26th International Conference on
  Computational Linguistics: Technical Papers}, 2016, pp. 2012--2021.

\bibitem{ji2016latent}
Y.~Ji, G.~Haffari, and J.~Eisenstein, ``A latent variable recurrent neural
  network for discourse relation language models,'' \emph{arXiv preprint
  arXiv:1603.01913}, 2016.

\bibitem{miyato2016adversarial}
T.~Miyato, A.~M. Dai, and I.~Goodfellow, ``Adversarial training methods for
  semi-supervised text classification,'' \emph{arXiv preprint
  arXiv:1605.07725}, 2016.

\bibitem{xiong2016dynamic}
C.~Xiong, S.~Merity, and R.~Socher, ``Dynamic memory networks for visual and
  textual question answering,'' \emph{international conference on machine
  learning}, pp. 2397--2406, 2016.

\bibitem{reithinger1997dialogue}
N.~Reithinger and M.~Klesen, ``Dialogue act classification using language
  models,'' in \emph{Fifth European Conference on Speech Communication and
  Technology}, 1997.

\bibitem{geertzen2007multidimensional}
J.~Geertzen, V.~Petukhova, and H.~Bunt, ``A multidimensional approach to
  utterance segmentation and dialogue act classification,'' in
  \emph{Proceedings of the 8th SIGdial Workshop on Discourse and Dialogue,
  Antwerp}, 2007, pp. 140--149.

\bibitem{surendran2006dialog}
D.~Surendran and G.-A. Levow, ``Dialog act tagging with support vector machines
  and hidden markov models,'' in \emph{Ninth International Conference on Spoken
  Language Processing}, 2006.

\bibitem{kalchbrenner2013recurrent}
N.~Kalchbrenner and P.~Blunsom, ``Recurrent convolutional neural networks for
  discourse compositionality,'' \emph{arXiv preprint arXiv:1306.3584}, 2013.

\bibitem{lee2016sequential}
J.~Y. Lee and F.~Dernoncourt, ``Sequential short-text classification with
  recurrent and convolutional neural networks,'' \emph{arXiv preprint
  arXiv:1603.03827}, 2016.

\bibitem{graves2014neural}
A.~Graves, G.~Wayne, and I.~Danihelka, ``Neural turing machines,'' \emph{arXiv
  preprint arXiv:1410.5401}, 2014.

\bibitem{bahdanau2014neural}
D.~Bahdanau, K.~Cho, and Y.~Bengio, ``Neural machine translation by jointly
  learning to align and translate,'' \emph{arXiv preprint arXiv:1409.0473},
  2014.

\bibitem{you2016image}
Q.~You, H.~Jin, Z.~Wang, C.~Fang, and J.~Luo, ``Image captioning with semantic
  attention,'' in \emph{Proceedings of the IEEE conference on computer vision
  and pattern recognition}, 2016, pp. 4651--4659.

\bibitem{wan2018improving}
Y.~Wan, Z.~Zhao, M.~Yang, G.~Xu, H.~Ying, J.~Wu, and P.~S. Yu, ``Improving
  automatic source code summarization via deep reinforcement learning,'' in
  \emph{Proceedings of the 33rd ACM/IEEE International Conference on Automated
  Software Engineering}.\hskip 1em plus 0.5em minus 0.4em\relax ACM, 2018, pp.
  397--407.

\bibitem{zhao2017video}
Z.~Zhao, Q.~Yang, D.~Cai, X.~He, and Y.~Zhuang, ``Video question answering via
  hierarchical spatio-temporal attention networks,'' in \emph{International
  Joint Conference on Artificial Intelligence (IJCAI)}, vol.~2, 2017, p.~1.

\bibitem{pan2017memen}
B.~Pan, H.~Li, Z.~Zhao, B.~Cao, D.~Cai, and X.~He, ``Memen: multi-layer
  embedding with memory networks for machine comprehension,'' \emph{arXiv
  preprint arXiv:1707.09098}, 2017.

\bibitem{luong2015effective}
M.-T. Luong, H.~Pham, and C.~D. Manning, ``Effective approaches to
  attention-based neural machine translation,'' \emph{arXiv preprint
  arXiv:1508.04025}, 2015.

\bibitem{kumar2016ask}
A.~Kumar, O.~Irsoy, P.~Ondruska, M.~Iyyer, J.~Bradbury, I.~Gulrajani, V.~Zhong,
  R.~Paulus, and R.~Socher, ``Ask me anything: Dynamic memory networks for
  natural language processing,'' in \emph{International Conference on Machine
  Learning}, 2016, pp. 1378--1387.

\bibitem{goodfellow2015explaining}
I.~J. Goodfellow, J.~Shlens, and C.~Szegedy, ``Explaining and harnessing
  adversarial examples,'' \emph{international conference on learning
  representations}, 2015.

\bibitem{tramer2017ensemble}
F.~Tram{\`e}r, A.~Kurakin, N.~Papernot, I.~Goodfellow, D.~Boneh, and
  P.~McDaniel, ``Ensemble adversarial training: Attacks and defenses,''
  \emph{ICLR}, 2018.

\bibitem{wang2018not}
J.~Wang, J.~Zhang, W.~Bao, X.~Zhu, B.~Cao, and P.~S. Yu, ``Not just privacy:
  Improving performance of private deep learning in mobile cloud,'' in
  \emph{Proceedings of the 24th ACM SIGKDD International Conference on
  Knowledge Discovery \& Data Mining}.\hskip 1em plus 0.5em minus 0.4em\relax
  ACM, 2018, pp. 2407--2416.

\bibitem{zhang2018layerwise}
J.~Zhang, J.~Wang, L.~He, Z.~Li, and P.~S. Yu, ``Layerwise perturbation-based
  adversarial training for hard drive health degree prediction,''
  \emph{International Conference on Data Mining}, 2018.

\bibitem{wu2017adversarial}
Y.~Wu, D.~Bamman, and S.~Russell, ``Adversarial training for relation
  extraction,'' in \emph{Proceedings of the 2017 Conference on Empirical
  Methods in Natural Language Processing}, 2017, pp. 1778--1783.

\bibitem{srivastava2014dropout}
N.~Srivastava, G.~Hinton, A.~Krizhevsky, I.~Sutskever, and R.~Salakhutdinov,
  ``Dropout: A simple way to prevent neural networks from overfitting,''
  \emph{The Journal of Machine Learning Research}, vol.~15, no.~1, pp.
  1929--1958, 2014.

\bibitem{iyyer2015deep}
M.~Iyyer, V.~Manjunatha, J.~Boyd-Graber, and H.~Daum{\'e}~III, ``Deep unordered
  composition rivals syntactic methods for text classification,'' in
  \emph{Proceedings of the 53rd Annual Meeting of the Association for
  Computational Linguistics and the 7th International Joint Conference on
  Natural Language Processing (Volume 1: Long Papers)}, vol.~1, 2015, pp.
  1681--1691.

\bibitem{xie2017data}
Z.~Xie, S.~I. Wang, J.~Li, D.~L{\'e}vy, A.~Nie, D.~Jurafsky, and A.~Y. Ng,
  ``Data noising as smoothing in neural network language models,'' \emph{arXiv
  preprint arXiv:1703.02573}, 2017.

\bibitem{li2017robust}
Y.~Li, T.~Cohn, and T.~Baldwin, ``Robust training under linguistic adversity,''
  in \emph{Proceedings of the 15th Conference of the European Chapter of the
  Association for Computational Linguistics: Volume 2, Short Papers}, vol.~2,
  2017, pp. 21--27.

\bibitem{Hochreiter:1997:LSM:1246443.1246450}
\BIBentryALTinterwordspacing
S.~Hochreiter and J.~Schmidhuber, ``Long short-term memory,'' \emph{Neural
  Comput.}, vol.~9, no.~8, pp. 1735--1780, Nov. 1997. [Online]. Available:
  \url{http://dx.doi.org/10.1162/neco.1997.9.8.1735}
\BIBentrySTDinterwordspacing

\bibitem{chung2014empirical}
J.~Chung, C.~Gulcehre, K.~Cho, and Y.~Bengio, ``Empirical evaluation of gated
  recurrent neural networks on sequence modeling,'' \emph{arXiv: Neural and
  Evolutionary Computing}, 2014.

\bibitem{pennington2014glove}
J.~Pennington, R.~Socher, and C.~Manning, ``Glove: Global vectors for word
  representation,'' in \emph{Proceedings of the 2014 conference on empirical
  methods in natural language processing (EMNLP)}, 2014, pp. 1532--1543.

\bibitem{kumar2017dialogue}
H.~Kumar, A.~Agarwal, R.~Dasgupta, S.~Joshi, and A.~Kumar, ``Dialogue act
  sequence labeling using hierarchical encoder with crf,'' \emph{arXiv preprint
  arXiv:1709.04250}, 2017.

\bibitem{chan2016listen}
W.~Chan, N.~Jaitly, Q.~Le, and O.~Vinyals, ``Listen, attend and spell: A neural
  network for large vocabulary conversational speech recognition,'' in
  \emph{Acoustics, Speech and Signal Processing (ICASSP), 2016 IEEE
  International Conference on}.\hskip 1em plus 0.5em minus 0.4em\relax IEEE,
  2016, pp. 4960--4964.

\bibitem{cho2014learning}
K.~Cho, B.~Van~Merri{\"e}nboer, C.~Gulcehre, D.~Bahdanau, F.~Bougares,
  H.~Schwenk, and Y.~Bengio, ``Learning phrase representations using rnn
  encoder-decoder for statistical machine translation,'' \emph{arXiv preprint
  arXiv:1406.1078}, 2014.

\bibitem{hochreiter1997long}
S.~Hochreiter and J.~Schmidhuber, ``Long short-term memory,'' \emph{Neural
  computation}, vol.~9, no.~8, pp. 1735--1780, 1997.

\bibitem{peng2015towards}
B.~Peng, Z.~Lu, H.~Li, and K.-F. Wong, ``Towards neural network-based
  reasoning,'' \emph{arXiv preprint arXiv:1508.05508}, 2015.

\bibitem{kingma2014adam}
D.~Kingma and J.~Ba, ``Adam: A method for stochastic optimization,''
  \emph{arXiv preprint arXiv:1412.6980}, 2014.

\bibitem{godfrey1992switchboard}
J.~J. Godfrey, E.~C. Holliman, and J.~McDaniel, ``Switchboard: Telephone speech
  corpus for research and development,'' in \emph{Acoustics, Speech, and Signal
  Processing, 1992. ICASSP-92., 1992 IEEE International Conference on},
  vol.~1.\hskip 1em plus 0.5em minus 0.4em\relax IEEE, 1992, pp. 517--520.

\bibitem{anderson1991hcrc}
A.~H. Anderson, M.~Bader, E.~G. Bard, E.~Boyle, G.~Doherty, S.~Garrod,
  S.~Isard, J.~Kowtko, J.~McAllister, J.~Miller \emph{et~al.}, ``The hcrc map
  task corpus,'' \emph{Language and speech}, vol.~34, no.~4, pp. 351--366,
  1991.

\bibitem{liu2017using}
Y.~Liu, K.~Han, Z.~Tan, and Y.~Lei, ``Using context information for dialog act
  classification in dnn framework,'' in \emph{Proceedings of the 2017
  Conference on Empirical Methods in Natural Language Processing}, 2017, pp.
  2170--2178.

\bibitem{tran2017aprocessings}
Q.~H. Tran, I.~Zukerman, and G.~Haffari, ``Preserving distributional
  information in dialogue act classification,'' \emph{Proceedings of the 2017
  Conference on Empirical Methods in Natural Language Processing}, pp.
  2151--2156, 2017.

\bibitem{ji2016a}
Y.~Ji, G.~Haffari, and J.~Eisenstein, ``A latent variable recurrent neural
  network for discourse relation language models,'' \emph{north american
  chapter of the association for computational linguistics}, pp. 332--342,
  2016.

\bibitem{julia2010dialog}
F.~N. Julia, K.~M. Iftekharuddin, and A.~U. ISLAM, ``Dialog act classification
  using acoustic and discourse information of maptask data,''
  \emph{International Journal of Computational Intelligence and Applications},
  vol.~9, no.~04, pp. 289--311, 2010.

\bibitem{tran2017bgenerative}
Q.~H. Tran, G.~Haffari, and I.~Zukerman, ``A generative attentional neural
  network model for dialogue act classification,'' in \emph{Proceedings of the
  55th Annual Meeting of the Association for Computational Linguistics (Volume
  2: Short Papers)}, vol.~2, 2017, pp. 524--529.

\end{thebibliography}

\end{document}